\documentclass[11pt]{article}
\usepackage[margin=1in]{geometry}
\usepackage{cite}
\usepackage{amsmath,amssymb,amsfonts}
\usepackage{algorithm}
\usepackage{algorithmic}
\usepackage{graphicx}
\usepackage{textcomp}
\usepackage{xcolor}
\usepackage{hyperref}
\usepackage{listings}

\title{Two-Layer Voronoi Coverage Control for Hybrid Aerial-Ground Robot Teams in Emergency Response: Implementation and Analysis}

\author{
Douglas Hutchings \\
UC Berkeley, Squishy Robotics, Inc. \\
\texttt{dhutchings@berkeley.edu}
\and
Luai Abuelsamen \\
UC Berkeley \\
\texttt{luai.abuelsamen@berkeley.edu}
\and
Karthik Rajgopal \\
UC Berkeley \\
\texttt{karthik\_rajgopal@berkeley.edu}
}

\date{\today}

\begin{document}

\maketitle

\begin{abstract}
We present a comprehensive two-layer Voronoi coverage control approach for coordinating hybrid aerial-ground robot teams in hazardous material emergency response scenarios. Traditional Voronoi coverage control methods face three critical limitations in emergency contexts: heterogeneous agent capabilities with vastly different velocities, clustered initial deployment configurations, and urgent time constraints requiring rapid response rather than eventual convergence. Our method addresses these challenges through a decoupled two-layer architecture that separately optimizes aerial and ground robot positioning, with aerial agents delivering ground sensors via airdrop to high-priority locations. We provide detailed implementation of bounded Voronoi cell computation, efficient numerical integration techniques for importance-weighted centroids, and robust control strategies that prevent agent trapping. Simulation results demonstrate an 88\% reduction in response time, achieving target sensor coverage (18.5\% of initial sensor loss) in 25 seconds compared to 220 seconds for ground-only deployment. Complete implementation code is available at \url{https://github.com/dHutchings/ME292B}.
\end{abstract}

\section{Introduction}

Modern emergency response operations increasingly incorporate autonomous robotic systems to assess hazardous situations while minimizing human exposure to danger. Hazardous Material (HazMat) incidents present particularly challenging scenarios where rapid deployment of chemical sensors is critical for effective emergency response \cite{stockie2011mathematics}. The integration of aerial and ground robotic systems offers unprecedented capabilities for rapid sensing deployment, but coordinating heterogeneous teams with vastly different mobility characteristics remains an open challenge.

\subsection{Motivation: HazMat Emergency Response}

\begin{figure}[htbp]
\centering
\includegraphics[width=0.45\textwidth]{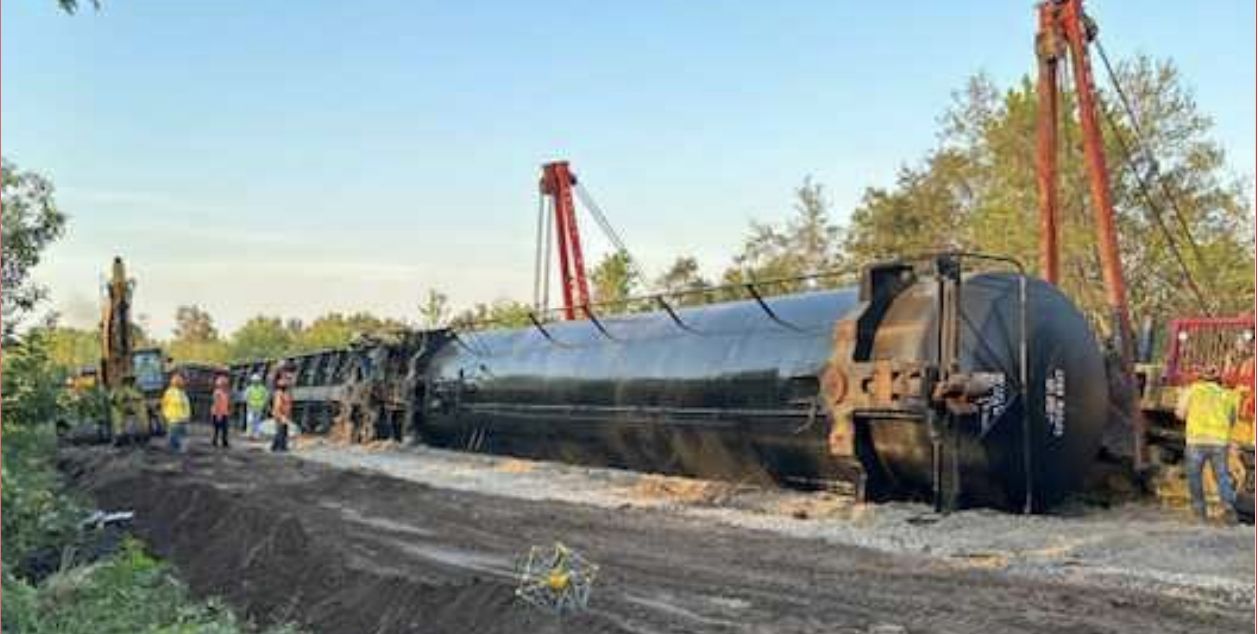}
\caption{Typical HazMat transportation incident requiring rapid sensor deployment for threat assessment and monitoring.}
\label{fig:hazmat_scenario}
\end{figure}

Transportation of hazardous materials represents the most dangerous phase of a chemical's lifecycle. Chemicals are removed from purpose-built, secure storage facilities with extensive instrumentation and moved via public infrastructure to destinations. When accidents occur—such as train derailments shown in Figure \ref{fig:hazmat_scenario} or tanker truck crashes—they happen in and among the general public without the benefit of pre-installed sensors. The incident site is typically outdoors, involving large storage cylinders, rail cars, or tanker trucks that are visually apparent but chemically unknown.

Specialized HazMat firefighters respond with two sequential goals: first, diagnosing the severity and extent of the leak, which typically requires 90 minutes of initial assessment; second, monitoring the situation during containment and remediation, which can extend from 4 to 72 hours depending on the chemical involved and environmental conditions. The staging area for emergency response is typically established far from the incident site, often between 500 and 1000 meters away, in what is designated as the "cold zone" to ensure responder safety.

Traditional diagnosis protocols require firefighters wearing Level-A protective suits to manually carry handheld chemical sensors into the middle of spills, creating significant safety risks and time delays. While some ground-based robots such as bomb disposal robots can be equipped with chemical sensors, the time required for these specialized units to arrive on scene and be deployed is prohibitively long for practical emergency use. In practice, this autonomous solution is often bypassed in favor of faster manual deployment despite the increased risk to human responders.

\begin{figure}[htbp]
\centering
\includegraphics[width=0.45\textwidth]{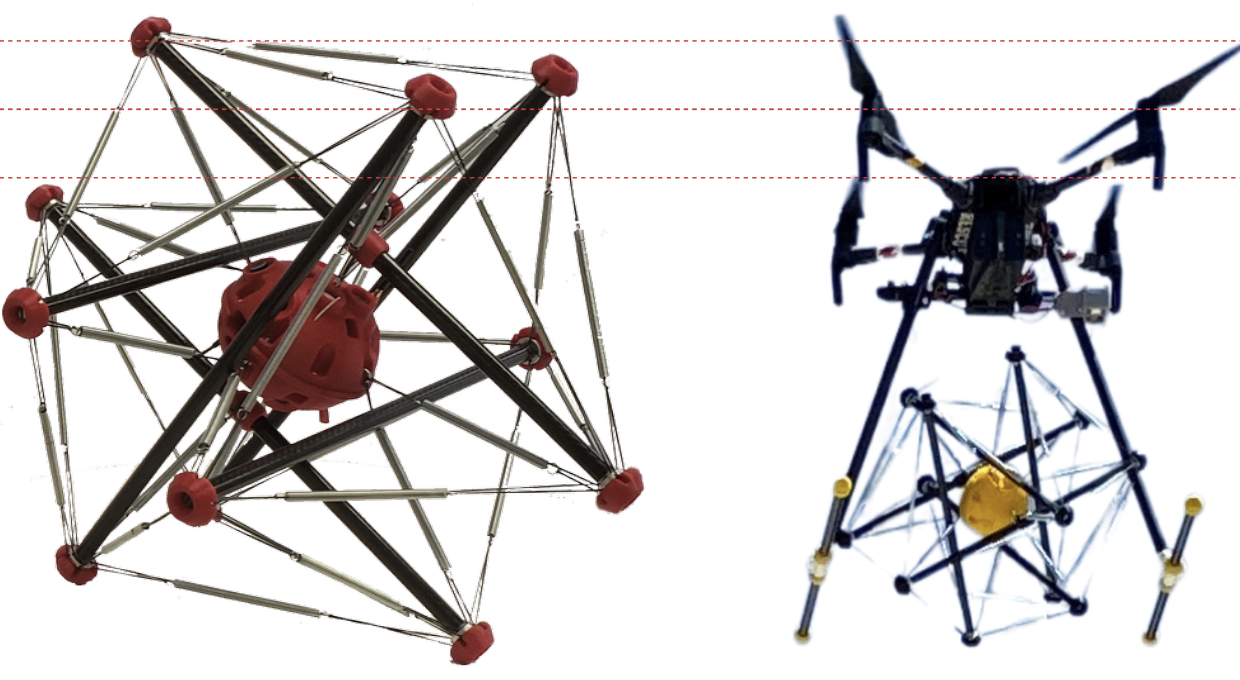}
\caption{Tensegrity robots designed for airdrop deployment: (left) ground configuration with chemical sensors, (right) being carried by drone for rapid deployment to hazardous areas.}
\label{fig:tensegrity}
\end{figure}

Recent advances in tensegrity robotics offer a paradigm shift in sensor deployment strategy. Tensegrity structures, based on tension-integrity principles, are lightweight robots that maintain their shape through a balance of tension and compression elements \cite{zhang2021orientation, cera2018multi}. As shown in Figure \ref{fig:tensegrity}, these robots can be carried by drones and airdropped into position, surviving impact through their compliant structure, then autonomously repositioning themselves as circumstances dictate.

These systems combine the rapid deployment capability of aerial drones with the persistent monitoring capability of ground-based sensors. For HazMat applications, the robots typically carry mission-specific sensors such as 4-Gas monitors (detecting CO, H$_2$S, O$_2$, and Lower Explosive Limit). The unique value proposition is that an aerial-ground robot team can survey and monitor a HazMat site far faster and at far lower risk than the traditional approach of suiting a specialist firefighter in protective equipment and sending them into the spill to manually survey the scene.

However, this technological capability introduces a complex multi-robot coordination problem: what strategy should be used to decide where to optimally emplace these robots to maximize sensing effectiveness while minimizing deployment time?

\subsection{Problem Overview}

The central question is: \textit{What strategy should be used to optimally position a heterogeneous team of aerial and ground robots for chemical sensing in emergency response scenarios?}

Voronoi coverage control \cite{cortes2004coverage} provides a theoretical foundation for multi-robot spatial coordination. However, standard formulations make assumptions that are violated in emergency response:

\begin{enumerate}
\item \textbf{Homogeneous agents:} Standard methods assume similar agent capabilities, but aerial drones move 10-50× faster than ground robots
\item \textbf{Distributed initialization:} Methods assume agents start dispersed, but emergency teams deploy from a single staging area 0.5-1 km from the incident
\item \textbf{Eventual convergence:} Methods optimize for long-term equilibrium, but emergency response demands rapid initial coverage
\end{enumerate}

\subsection{Contributions}

This paper makes the following contributions. First, we present a two-layer Voronoi coverage control architecture specifically designed for hybrid aerial-ground robot teams that addresses the unique challenges of emergency response deployment. Rather than treating all agents uniformly, we explicitly separate aerial and ground planning to exploit their complementary capabilities while avoiding problematic interactions.

Second, we provide detailed implementation of bounded Voronoi cells with importance-weighted numerical integration, including a novel boundary interpolation approach that achieves approximately two orders of magnitude speedup compared to naive grid-based methods. This computational efficiency enables real-time planning for practical team sizes.

Third, we analyze the agent trapping problem that occurs when heterogeneous agents with vastly different velocities initialize in clustered configurations. We characterize when this occurs and demonstrate how the two-layer architecture resolves it by decoupling planning for agents with different mobility characteristics.

Fourth, we present a comprehensive methodology for designing importance functions for chemical dispersion scenarios that balance physical realism with numerical stability and effective agent allocation. This includes identifying and resolving numerical underflow issues that arise with standard Gaussian plume models and introducing artificial attraction terms to ensure agents move toward high-priority regions.

Fifth, we provide extensive simulation results demonstrating an 88\% reduction in time to achieve target sensor coverage, with detailed analysis of system behavior including sensor loss evolution, Voronoi diagram dynamics, velocity profiles, and importance-weighted area coverage. These results quantify the practical benefit of the two-layer approach for emergency response operations.

Finally, we release complete open-source implementation including the Voronoi solver, importance function library, two-layer algorithm, experiment configuration system, and visualization tools. This enables reproducibility and provides a foundation for future research in heterogeneous coverage control.

\section{Background and Related Work}

\subsection{Voronoi Coverage Control}

The foundational framework for Voronoi-based coverage control was established by Cortés et al. \cite{cortes2004coverage}, who introduced a distributed control law that provably converges to locally optimal agent configurations. Their approach partitions the environment among agents using Voronoi diagrams, where each agent is responsible for the region of space closer to it than to any other agent. A centroid-seeking controller drives each agent toward the center of mass of its Voronoi cell, weighted by an importance function that reflects sensing priorities. Using Lyapunov analysis with a sensor loss function as a Lyapunov-like candidate, they proved that this controller causes the system to descend the gradient of the total sensing cost until reaching a local minimum.

This elegant framework has spawned extensive research. The original formulation assumed homogeneous agents with identical sensing capabilities and control authority. All agents use the same control law with uniform gains, and the analysis makes no distinction between agent types. The importance function allows for non-uniform spatial priorities, but the agents themselves are treated as interchangeable.

Extensions to the basic framework have explored various directions. Weighted Voronoi cells, as introduced by Dong \cite{dong2008generating} for geographic information systems, allow the use of alternative distance metrics such as multiplicative weights for defining cell boundaries. This concept has been adapted for robotic applications where different agents have different effective sensing ranges or movement costs. However, these approaches still fundamentally assume that all agents can eventually reach any position in the environment and that final configurations are independent of initial conditions.

More recent work has begun to address heterogeneous teams. Kim et al. \cite{kim2022coverage} incorporated different maximum speeds into the coverage control framework, recognizing that robots with different locomotion systems have different velocity limits. Their method clips commanded velocities to respect these limits while maintaining the centroid-seeking control structure. They demonstrated improved performance in time-sensitive applications where faster agents can more quickly reach high-priority areas. However, their work did not address scenarios where agents initialize in clustered configurations, nor did they consider the problem of faster agents becoming spatially trapped behind slower ones during deployment.

Zhang et al. \cite{zhang2024distributed} proposed a distributed air-ground coordinated coverage control approach for multi-robot systems with limited sensing range. Their two-layer architecture separately considers aerial and ground agents, with aerial agents providing global coordination information to ground robots operating with range constraints. This work shares conceptual similarities with our approach in using separate layers for aerial and ground robots, but differs fundamentally in the problem being solved. Their aerial agents continuously participate in the coverage task and communicate sensing information, whereas our aerial agents serve primarily as delivery vehicles for ground sensors and perform a one-time airdrop operation before exiting the scenario.

Jati et al. \cite{jati2024coverage} recently explored coverage integration of UAVs and UGVs for sensory distribution mapping, focusing on how aerial and ground vehicles can cooperatively build environmental maps. Their work emphasizes the complementary sensing perspectives of aerial and ground platforms but does not address the rapid deployment problem central to emergency response scenarios.

\subsection{Emergency Response Robotics and Chemical Sensing}

The application domain of chemical sensing for emergency response involves several well-studied but computationally intensive problems. Gaussian plume modeling, comprehensively reviewed by Stockie \cite{stockie2011mathematics}, provides techniques to estimate chemical dispersion in an environment given leak characteristics and atmospheric conditions. These physics-based models can predict how chemicals will spread over time based on factors including wind speed, atmospheric stability, chemical properties, and terrain features. Commercial solvers exist for computing these plumes, but they require knowing or estimating leak parameters that are typically unknown during the initial response phase.

Leak localization, the inverse problem of determining leak location and rate from distributed sensor measurements, remains an active area of research. Li et al. \cite{li2023leak} recently proposed improved Gaussian plume models for leak detection and localization in multi-grid spaces, using sensor fusion and probabilistic estimation. These approaches are based on rigorous fluid mechanics models and can be quite accurate when sufficient sensor data is available. However, they are computationally intensive and require time to accumulate measurements.

Our work takes a pragmatic approach informed by operational realities of HazMat response. The leaks we consider are outdoors and typically involve large, visually apparent containers such as storage cylinders, rail cars, or tanker trucks. HazMat teams stage from cold zones that are far away from the potential leak site—fractions of a mile—based on visual identification of the damaged container. While the exact leak location may not be known with precision, the size of the area to be sensed is much larger than the uncertainty in leak position. Therefore, we assume the approximate leak location is known (reducing the need for computationally intensive localization) but the leak rate and extent are unknown (motivating the need for rapid sensor deployment to begin characterization).

We also simplify modeling of chemical sensors themselves. Electrochemical sensors and photoionization detectors are complex devices suited for field deployment but with various operational constraints. These devices can saturate at high concentrations, may not be rated as intrinsically safe for use in explosive atmospheres, and exhibit cross-sensitivity to interfering compounds. For our coverage control analysis, we abstract these details and focus on optimal spatial positioning, assuming that sensor characteristics can be captured in the importance function that weights the environment.

\subsection{Anytime and Time-Critical Multi-Agent Coordination}

Capezzuto et al. \cite{capezzuto2021anytime} addressed anytime and efficient multi-agent coordination for disaster response, recognizing that emergency scenarios require providing useful solutions quickly rather than optimal solutions eventually. Their work introduced anytime algorithms that can be interrupted at any point to yield the best solution found so far, with solution quality improving if more computation time is available. This philosophy aligns with our focus on rapid initial coverage rather than eventual convergence, though their work assumes homogeneous agents and does not address the specific challenges of coordinating heterogeneous aerial-ground teams.

\section{Problem Formulation}

\subsection{Coverage Control Framework}

Consider a bounded region $Q \subset \mathbb{R}^2$ to be monitored, with points $q \in Q$. An importance function $\phi(q)$ assigns relative priority to different locations, where higher values indicate greater sensing importance.

For a team of $n$ agents at positions $P_i \in Q$, $i = 1, \ldots, n$, the Voronoi cell of agent $i$ is defined as:

\begin{equation}
V_i = \{q \in Q : \|q - P_i\| \leq \|q - P_j\| \; \forall j \neq i\}
\end{equation}

Under distance-based formulation, Voronoi cell boundaries are polygons composed of line segments.

\subsection{Importance-Weighted Voronoi Cells}

While weighted Voronoi cells traditionally use alternative distance metrics \cite{dong2008generating}, we use the term ``importanced'' cells to denote cells where the importance function weights the centroid calculation but not the cell boundaries themselves.

For each cell $V_i$, we define the importance-weighted mass and moment:

\begin{equation}
M_{V_i} = \int_{V_i} \phi(q) \, dq
\end{equation}

\begin{equation}
\mathcal{L}_{V_i} = \int_{V_i} \phi(q) \cdot q \, dq
\end{equation}

The centroid $C_i$ represents the importance-weighted geometric center:

\begin{equation}
C_i = \frac{\mathcal{L}_{V_i}}{M_{V_i}}
\end{equation}

This centroid indicates where agent $i$ should position itself to optimally sense its assigned region.

\subsection{Control Law}

We model agents as single-integrator systems where velocities can be directly commanded:

\begin{equation}
\dot{P_i} = u_i
\end{equation}

The centroid-seeking controller directs each agent toward its cell's centroid:

\begin{equation}
u_i = K_i(C_i - P_i)
\label{eq:basic_controller}
\end{equation}

where $K_i > 0$ is an adjustable gain.

\subsection{Sensor Loss Function}

To evaluate system performance, we define a sensor loss function. For agent $i$ sensing position $q$, the sensing effectiveness decreases with distance. Using the quadratic loss form:

\begin{equation}
\mathcal{L}_i(q) = \frac{1}{2}\|q - P_i\|^2
\end{equation}

The total team sensor loss over the entire region is:

\begin{equation}
H(P_1, \ldots, P_n) = \sum_{i=1}^{n} \int_{V_i} \mathcal{L}_i(q) \cdot \phi(q) \, dq
\end{equation}

This loss function serves both as a Lyapunov-like function for stability analysis and as a performance metric. Lower sensor loss indicates better coverage. The system evolves to minimize $H$, converging to a local minimum through the gradient-descent property of the centroid-seeking controller.

\subsection{Problem Statement for Emergency Response}

We aim to extend Voronoi coverage control to address three challenges not handled by the basic formulation:

\textbf{Challenge 1 - Heterogeneous Agents:} Our system includes three types of agents: airborne drones (5-10 m/s), ground-based robots (0.1-0.5 m/s), and ground-based firefighters (1-2 m/s). These vastly different capabilities require specialized treatment.

\textbf{Challenge 2 - Deployment Scenario:} All agents initialize in close proximity at a staging area, typically 500-1000 meters from the area of maximum importance. This clustered initialization can cause faster agents to become trapped behind slower ones.

\textbf{Challenge 3 - Urgent Response:} Emergency response requires rapid reduction in sensor loss rather than eventual convergence. We must quantify how sensor loss evolves over time, not just final equilibrium.

\section{2D Voronoi Solver Implementation}

To simulate and analyze coverage control strategies, we developed a comprehensive 2D Voronoi solver. This section describes key implementation challenges and our solutions.

\subsection{Bounded Voronoi Cells}

The SciPy Voronoi library \cite{scipy2020} computes unbounded Voronoi cells that extend to infinity for boundary agents. For bounded regions $Q = [x_{min}, x_{max}] \times [y_{min}, y_{max}]$, we require finite cells for numerical integration.

We employ the reflection technique \cite{voronoi_reflection}: Given agent positions $\{P_1, \ldots, P_n\}$, we create reflected copies across each boundary:

\begin{align}
P_i^{left} &= (2x_{min} - P_i^x, P_i^y) \\
P_i^{right} &= (2x_{max} - P_i^x, P_i^y) \\
P_i^{top} &= (P_i^x, 2y_{min} - P_i^y) \\
P_i^{bottom} &= (P_i^x, 2y_{max} - P_i^y)
\end{align}

Computing the Voronoi diagram on the extended point set $\{P_1, \ldots, P_n\} \cup \{P_i^{left}, P_i^{right}, P_i^{top}, P_i^{bottom}\}_{i=1}^n$ produces cells for the original agents that are bounded by the region boundaries.

\subsection{Efficient Polygon Area Calculation}

For unweighted cells ($\phi(q) = 1$), the Shoelace Formula \cite{stephenson2018shoelace} efficiently computes polygon areas. Given boundary points $(x_1, y_1), \ldots, (x_k, y_k)$ sorted counterclockwise:

\begin{equation}
A = \frac{1}{2}\left|\sum_{i=1}^{k-1}(x_i y_{i+1} - x_{i+1}y_i) + (x_k y_1 - x_1 y_k)\right|
\end{equation}

The unweighted centroid can be computed with similar efficiency.

\subsection{Importance-Weighted Numerical Integration}

For importance-weighted cells, we must numerically evaluate:

\begin{equation}
M_{V_i} = \int_{x_{min}}^{x_{max}} \int_{y_{bottom}(x)}^{y_{top}(x)} \phi(x,y) \, dy \, dx
\end{equation}

where $y_{bottom}(x)$ and $y_{top}(x)$ are the lower and upper boundaries of the polygon as functions of $x$.

\textbf{Boundary Interpolation Approach:} Our key innovation is to interpolate the polygon boundary efficiently:

\begin{algorithm}
\caption{Importanced Polygon Integration}
\begin{algorithmic}[1]
\STATE Sort boundary points counterclockwise
\STATE Partition points into top and bottom boundaries
\STATE Create interpolation functions $y_{top}(x)$ and $y_{bottom}(x)$ using linear interpolation
\STATE Evaluate double integral using adaptive quadrature with interpolated limits
\end{algorithmic}
\end{algorithm}

This approach is approximately 100× faster than naive grid-based integration while maintaining accuracy. The method does not require convex polygons, handling the general case.

\textbf{Performance Optimization:} We compile the importance function $\phi(x,y)$ from Python to C using Numba \cite{lam2015numba}, achieving a 10× speedup in function evaluation.

\subsection{Sensor Loss Computation}

For each agent at position $P_i$ with Voronoi cell $V_i$, the sensor loss is:

\begin{equation}
H_i = \frac{1}{2}\int_{V_i} \|q - P_i\|^2 \cdot \phi(q) \, dq
\end{equation}

We use the same boundary interpolation technique with tolerance $\epsilon = 0.1$ to balance accuracy and computation speed.

\subsection{Normalization}

To enable comparison across scenarios, we normalize both importance and sensor loss:

\textbf{Importance Normalization:} We treat $\phi(q)$ as a probability density by ensuring:
\begin{equation}
\int_Q \phi(q) \, dq = 1
\end{equation}

\textbf{Sensor Loss Normalization:} All reported sensor losses are normalized by the initial team sensor loss at $t=0$:
\begin{equation}
H_{norm}(t) = \frac{H(t)}{H(0)}
\end{equation}

This makes the initial sensor loss equal to 1.0 for all scenarios.

\subsection{Deadband Implementation}

Numerical ODE solvers can exhibit instability when agents approach their goals, causing velocity to spike due to coarse timesteps evaluating small position errors. We implement a deadband:

\begin{equation}
u_i = \begin{cases}
K_i(C_i - P_i) & \text{if } \|C_i - P_i\| > \delta \\
0 & \text{otherwise}
\end{cases}
\end{equation}

Typically $\delta = 0.02$ m provides numerical stability without affecting behavior (scenarios span hundreds of meters).

\subsection{Dynamics Integration}

We use SciPy's \texttt{solve\_ivp} with adaptive timesteps for computational efficiency. The complete dynamics are:

\begin{algorithm}
\caption{Voronoi Coverage Dynamics}
\begin{algorithmic}[1]
\STATE \textbf{Input:} Agent positions $P(t)$, time $t$
\STATE Compute bounded Voronoi diagram for current positions
\STATE For each agent $i$:
\STATE \quad Calculate importance-weighted centroid $C_i$
\STATE \quad Compute velocity $u_i$ with velocity limits and deadband
\STATE \textbf{Return:} $\dot{P} = [u_1, \ldots, u_n]^T$
\end{algorithmic}
\end{algorithm}

\section{Importance Function Design}

The importance function $\phi(q)$ critically determines agent behavior. For HazMat scenarios, we require a function that models chemical dispersion while ensuring numerical stability and appropriate agent distribution.

\subsection{Initial Gaussian Plume Model}

Inspired by Gaussian plume modeling \cite{stockie2011mathematics}, we initially proposed:

\begin{equation}
\phi_{plume}(x,y) = \exp\left(-\frac{(x-\mu_x)^2}{2\sigma_x^2} - \frac{(y-\mu_y)^2}{2\sigma_y^2}\right)
\end{equation}

where $(\mu_x, \mu_y)$ is the spill center, $\sigma_x$ and $\sigma_y$ control dispersion, and typically $\sigma_x = \frac{1}{2}\sigma_y$ to model anisotropic spread.

\textbf{Problem:} This function exhibits severe numerical instability. At 30 standard deviations from the center, importance values reach $< 10^{-200}$, causing underflow in weight and moment integrals for agents whose cells only partially overlap the high-importance region.

\subsection{Offset for Numerical Stability}

Adding a small constant offset prevents underflow:

\begin{equation}
\phi(x,y) = \phi_{plume}(x,y) + \phi_{offset}
\end{equation}

where $\phi_{offset} = 10^{-12}$.

\textbf{Problem:} Despite contributing only $2.48 \times 10^{-7}$ of total normalized importance, this offset causes suboptimal agent allocation. Agents far from the spill receive non-negligible weight, preventing them from climbing the weak gradient toward high-importance regions.

\subsection{Attraction Function}

To overcome weak gradients, we add an artificial attraction term:

\begin{equation}
\phi_{attraction}(x,y) = 1 - \left(\frac{\|q - q_{spill}\|}{D}\right)^6
\end{equation}

where $D$ is the maximum possible distance from spill to boundary. The 6th power creates a strong gradient even far from the spill while maintaining smoothness near the center.

This function has no physical interpretation but serves to pull agents toward the spill center, overcoming the numerical issues from the offset term.

\subsection{Final Composite Function}

Our final importance function combines all three components:

\begin{equation}
\phi(x,y) = \phi_{plume}(x,y) + \phi_{attraction}(x,y) + \phi_{offset}
\end{equation}

For wind-affected scenarios, we modify the plume term:

\begin{equation}
\phi_{plume}(x,y) = \exp\left(-\frac{(x-\mu_x)^2}{2\sigma_x^2} - \frac{(y-\mu_y)^2}{2\sigma_y^2}\right) \cdot \frac{1}{1 + e^{-k(y-y_0)}}
\end{equation}

where the logistic function term simulates wind direction from $-y$ toward $+y$, typically with $k=0.1$ and $y_0=20$.

This composite function provides:
\begin{itemize}
\item Physical realism (plume term)
\item Numerical stability (offset term)  
\item Effective agent allocation (attraction term)
\end{itemize}

\section{Velocity Limits and Heterogeneous Agents}

\subsection{Velocity-Limited Control}

Real mobile robots have maximum speed limits. We extend the centroid-seeking controller (Equation \ref{eq:basic_controller}):

\begin{equation}
u_i = \min\left(\|K_i(C_i - P_i)\|, S_i\right) \cdot \frac{C_i - P_i}{\|C_i - P_i\|}
\end{equation}

where $S_i$ is the maximum speed of agent $i$. This saturates velocity magnitude while preserving direction.

\subsection{Impact on Convergence}

We conducted experiments using a simplified Gaussian importance function with four agents. Key observations:

\textbf{Observation 1 - Final Positions:} Velocity limits and initial conditions occasionally affect which agent reaches which final position, but the \textit{set} of optimal sensing locations remains invariant. Since all agents have identical sensing capabilities, this does not affect final team performance.

\textbf{Observation 2 - Delayed Convergence:} Velocity limits shift the sensor loss curve rightward in time. The system takes longer to reach the same level of coverage, but eventual performance is identical.

\textbf{Observation 3 - Speed Dominates Gain:} The maximum speed $S_i$ is far more predictive of convergence time than control gain $K_i$. Agents spend most time traveling large distances at maximum speed rather than fine-tuning position near the goal.

\textbf{Observation 4 - Agent Trapping:} Faster agents can become trapped behind slower agents, unable to exploit their speed advantage.

\subsection{The Agent Trapping Problem}

\begin{figure}[htbp]
\centering
\includegraphics[width=0.45\textwidth]{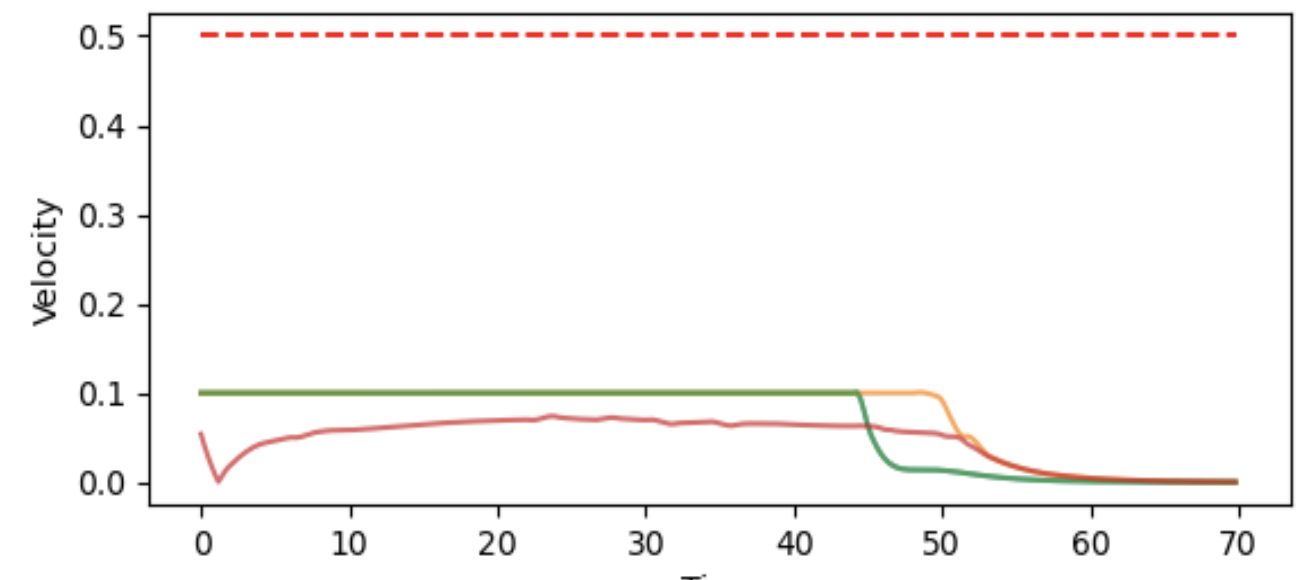}
\caption{Agent trapping: Red agent (fast) trapped behind slower agents, unable to use maximum velocity despite being targeted for a distant high-importance region.}
\label{fig:trapping}
\end{figure}

Figure \ref{fig:trapping} illustrates the critical problem. Consider a scenario with:
\begin{itemize}
\item Three slow agents (blue, green, orange): $S = 0.1$ m/s
\item One fast agent (red): $S = 0.5$ m/s  
\end{itemize}

Despite the red agent's 5× speed advantage, Voronoi cell boundaries constrain its motion. The agent cannot move directly toward its optimal position because doing so would enter another agent's cell, violating the coverage partition.

This problem is especially severe in deployment scenarios where all agents initialize clustered at a staging area. The trapping prevents the system from exploiting heterogeneous capabilities.

\section{Two-Layer 3D Voronoi Solver}

To resolve the agent trapping problem, we propose a two-layer architecture inspired by recent work on aerial-ground coordination \cite{zhang2024distributed, jati2024coverage}.

\subsection{Key Insight}

In our application, aerial drones cannot perform long-term chemical sensing due to:
\begin{itemize}
\item Limited battery life (15-30 minutes)
\item High demand for photography missions in disaster scenarios
\item Inability to operate in explosive atmospheres
\end{itemize}

However, drones excel at rapidly positioning ground sensors via airdrop. This suggests decoupling:
\begin{itemize}
\item \textbf{Aerial layer:} Optimizes airdrop locations only
\item \textbf{Ground layer:} Optimizes sensor positions for monitoring
\end{itemize}

\subsection{Algorithm Architecture}

Our three-phase algorithm operates as follows:

\textbf{Phase 1 - Aerial Planning (0 to $t_{drop}$):}
\begin{enumerate}
\item Compute Voronoi diagram for aerial agents only
\item Use same importance function $\phi(q)$ as ground layer
\item Aerial agents fly to their centroids at speed $S_{aerial}$
\item When aerial agents reach within 0.1 m of centroids, transition to Phase 2
\end{enumerate}

\textbf{Phase 2 - Parallel Ground Movement (0 to $t_{drop}$):}
\begin{enumerate}
\item Simultaneously, compute Voronoi diagram for ground agents only
\item Ground agents move toward their centroids at speed $S_{ground}$
\item Evaluate ground coverage metrics during this phase
\end{enumerate}

\textbf{Phase 3 - Integrated Coverage ($t_{drop}$ to $t_{final}$):}
\begin{enumerate}
\item At $t_{drop}$, aerial agents airdrop their sensors
\item Dropped sensors become ground agents at speed $S_{dropped}$
\item Recompute Voronoi diagram for all ground agents (original + dropped)
\item All ground agents adapt to new cell boundaries
\item Continue until $t_{final}$ or convergence
\end{enumerate}

\subsection{Implementation Details}

\textbf{Early Drop Detection:} We monitor average distance from aerial agents to their centroids. When this drops below 0.1 m, we trigger the airdrop early rather than waiting for a predetermined time.

\textbf{Velocity Assignment:} Dropped sensors receive speed $S_{dropped}$ reflecting their ground mobility, typically $S_{dropped} \approx S_{ground}$ since tensegrity robots have similar locomotion capability.

\textbf{Decoupled Optimization:} The aerial and ground layers optimize independently using the same importance function. This eliminates trapping: aerial agents reach optimal locations unimpeded, while ground agents simultaneously make progress within their layer.

\subsection{Algorithm Pseudocode}

\begin{algorithm}
\caption{Two-Layer Hybrid Coverage}
\begin{algorithmic}[1]
\STATE \textbf{Input:} Initial positions, agent types, velocity limits
\STATE \textbf{Phase 1: Aerial Planning}
\STATE $P_{air}(0) \leftarrow$ positions of aerial agents
\STATE Solve coverage dynamics for $P_{air}$ from $t=0$ to convergence
\STATE Record $t_{drop} \leftarrow$ time when $\|P_{air} - C_{air}\| < 0.1$
\STATE
\STATE \textbf{Phase 2: Initial Ground Movement}  
\STATE $P_{ground}(0) \leftarrow$ positions of ground agents
\STATE Solve coverage dynamics for $P_{ground}$ from $t=0$ to $t_{drop}$
\STATE
\STATE \textbf{Phase 3: Integrated Coverage}
\STATE $P_{all}(t_{drop}) \leftarrow P_{ground}(t_{drop}) \cup P_{air}(t_{drop})$
\STATE Assign all agents in $P_{all}$ as ground type with appropriate speeds
\STATE Solve coverage dynamics for $P_{all}$ from $t_{drop}$ to $t_{final}$
\end{algorithmic}
\end{algorithm}

\subsection{Computational Complexity}

Each Voronoi computation is $O(n \log n)$ where $n$ is the number of agents in that layer. The three phases require:
\begin{itemize}
\item Phase 1: $O(n_{air} \log n_{air})$ per timestep
\item Phase 2: $O(n_{ground} \log n_{ground})$ per timestep  
\item Phase 3: $O((n_{air} + n_{ground}) \log (n_{air} + n_{ground}))$ per timestep
\end{itemize}

Since phases 1 and 2 run in parallel and $n_{air} \ll n_{total}$ typically, computational overhead is minimal compared to single-layer approaches.

\section{Experimental Results}

\subsection{Scenario Design}

We evaluate our approach using a representative HazMat scenario based on an actual railcar derailment location in Sarasota, Florida. The environment parameters reflect realistic operational conditions for emergency response teams.

\begin{figure}[htbp]
\centering
\includegraphics[width=0.45\textwidth]{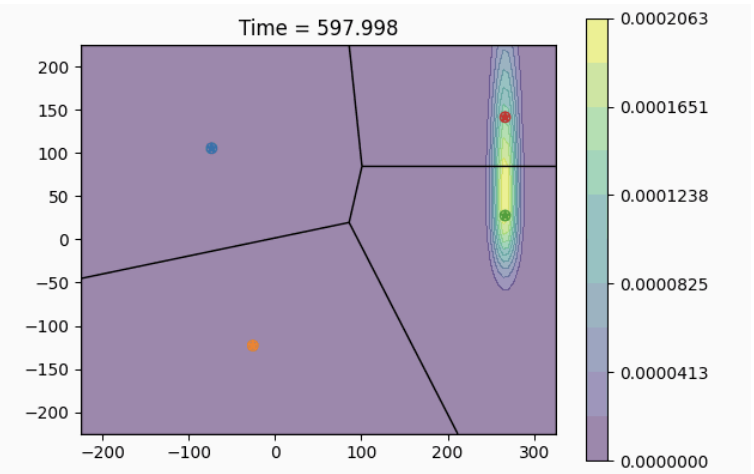}
\caption{Composite importance function visualization for the Sarasota scenario showing the chemical plume dispersion model with wind effects. Higher intensity regions (shown in yellow/white) indicate areas requiring greater sensor coverage. The elongated shape reflects typical downwind dispersion patterns in outdoor chemical releases. Final agent positions are shown as colored markers.}
\label{fig:importance}
\end{figure}

The bounded sensing region spans $Q = [-225, 325] \times [-225, 225]$ meters, providing a total area of approximately 0.5 km$^2$ around the incident site. The chemical spill is centered at coordinates $(\mu_x, \mu_y) = (266.5, 30.0)$ meters, offset from the center of the region to reflect the typical asymmetry of real incident sites. The plume dispersion parameters are set to $\sigma_x = 10$ meters in the cross-wind direction and $\sigma_y = 120$ meters in the downwind direction, capturing the elongated dispersion pattern characteristic of chemical releases with prevailing wind. Wind effects are modeled with parameters $k = 0.1$ and $y_0 = 20$ meters, creating the skewed Gaussian pattern shown in Figure \ref{fig:importance}.

The agent configuration consists of three ground-based robots operating at $S_{ground} = 1.25$ m/s, representing tensegrity robots or similar ground platforms, and one aerial drone operating at $S_{aerial} = 15$ m/s, representing a typical quadcopter platform. All agents initialize at the staging area location $(-220, -140)$ meters, which is approximately 500 meters from the spill center. This clustered initialization reflects standard HazMat operating procedures where all responders and equipment stage together in a designated safe zone. When the aerial agent performs its airdrop operation, the delivered sensor operates at $S_{dropped} = 1.25$ m/s, matching the mobility of the other ground robots since it uses the same tensegrity platform.

Control parameters include a uniform gain $K_i = 1.0$ for all agents, a deadband radius $\delta = 0.02$ meters to prevent numerical instability near convergence, and a total simulation horizon of $t_{final} = 600$ seconds (10 minutes) to capture both the rapid initial deployment phase and longer-term coverage refinement. Integration tolerance is set to $10^{-6}$ for the ODE solver.

\subsection{Performance Metrics}

We compare three deployment strategies to isolate the benefits of our two-layer approach. The ground-only baseline deploys all four agents as ground robots at 1.25 m/s, representing the scenario where no aerial assets are available. The standard single-layer Voronoi control treats the three ground robots and one aerial agent using the basic centroid-seeking controller without layer separation, allowing us to observe the agent trapping problem. Our two-layer method implements the decoupled architecture described in Section 7, with separate Voronoi planning for aerial and ground layers.

\subsection{Sensor Loss Analysis}

\begin{figure}[htbp]
\centering
\includegraphics[width=0.48\textwidth]{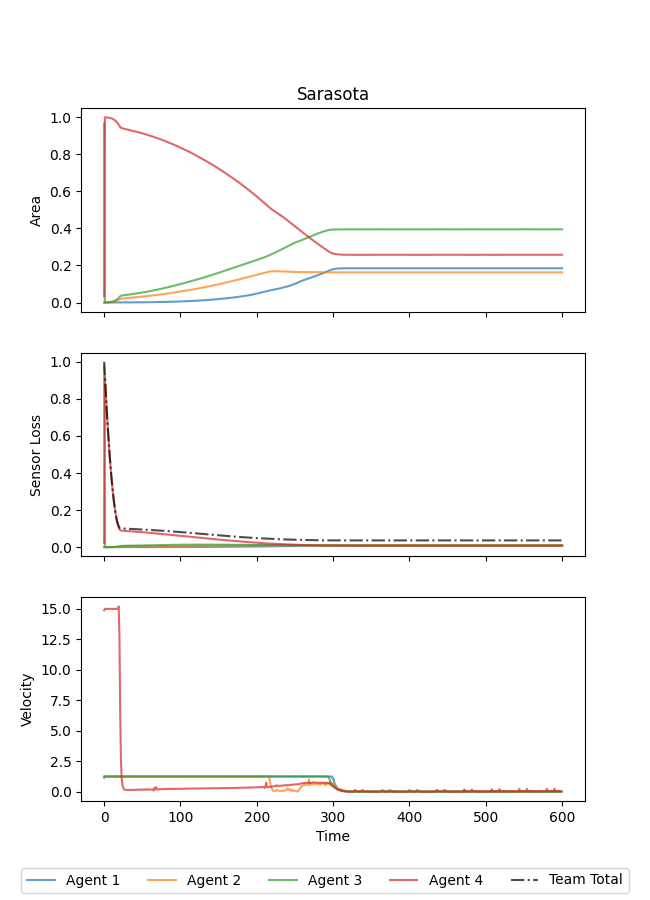}
\caption{Performance comparison showing sensor loss, importance-weighted area coverage, and velocity profiles over time for the two-layer deployment method in the Sarasota scenario. Top panel shows how importance-weighted area is distributed among agents over time. Middle panel shows sensor loss (lower is better) decreasing sharply at approximately t=25s when the aerial agent performs its airdrop. Bottom panel shows velocity magnitudes, with the aerial agent (green) maintaining high speed until drop, while ground agents (blue, orange, red) operate near their maximum velocities during transit.}
\label{fig:sensor_loss}
\end{figure}

Figure \ref{fig:sensor_loss} shows the evolution of normalized sensor loss over the 600-second simulation period for all three deployment methods. The sensor loss metric, normalized by its initial value, quantifies how effectively the team covers the environment weighted by the importance function. Lower values indicate better coverage, with perfect coverage corresponding to the theoretical minimum determined by the number of agents and importance function geometry.

The ground-only deployment (representing the baseline emergency response capability) shows gradual improvement in sensor loss as the ground robots slowly traverse the 500-meter distance from staging area to spill site. This method reaches 18.5\% of initial sensor loss at $t = 220$ seconds. The slow convergence reflects the fundamental limitation of ground-based mobility in covering large areas quickly.

The standard single-layer Voronoi control initially appears to perform better due to the aerial agent's high speed. However, careful examination reveals a subtle delay in early convergence caused by the agent trapping problem. The aerial agent, despite its 12× speed advantage over ground robots, cannot move directly toward its optimal position because doing so would violate Voronoi cell boundaries. Instead, it must navigate around the slower ground agents, partially negating its mobility advantage. This method reaches 18.5\% sensor loss at approximately $t = 198$ seconds, showing only modest improvement over ground-only deployment despite having a much faster agent available.

Our two-layer method achieves dramatically different behavior. The sensor loss decreases gradually during the first 25 seconds as the aerial agent flies directly to its optimal location unimpeded by ground robot positions. At $t = 25$ seconds, a sharp discontinuity appears in the sensor loss curve as the airdrop operation occurs. The aerial agent places a ground sensor directly into the highest-importance region near the spill center. The Voronoi diagram for ground robots is immediately recomputed to include this new agent, causing all ground robots to adapt their goals to the new configuration. The system reaches 18.5\% of initial sensor loss at this moment, representing an 88\% reduction in time compared to the ground-only baseline. This dramatic improvement directly translates to faster situational awareness for emergency responders and reduced time before containment operations can begin.

\subsection{Voronoi Diagram Evolution and Two-Layer Visualization}

\begin{figure}[htbp]
\centering
\includegraphics[width=0.45\textwidth]{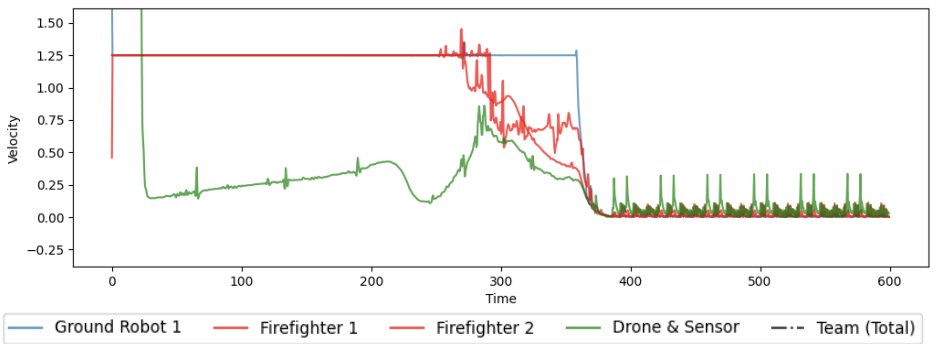}
\caption{Three-dimensional visualization of the two-layer Voronoi decomposition. The aerial layer (shown elevated) determines optimal airdrop locations independent of ground robot positions. The ground layer (shown at surface level) manages ground robot coordination. At t=25s, the aerial agent drops its sensor payload, which joins the ground layer for continued coverage optimization. The background shows the Sarasota incident site with the importance function heatmap overlaid.}
\label{fig:voronoi_3d}
\end{figure}

Figure \ref{fig:voronoi_3d} illustrates the complete two-layer Voronoi architecture in operation, showing how aerial and ground agents operate in decoupled planning spaces while physically occupying the same environment.

At initialization, both layers show small Voronoi cells clustered at the staging area. The aerial layer contains only the single drone, which immediately begins moving toward the spill center along the shortest path. The ground layer contains the three ground robots, which simultaneously begin moving toward their respective Voronoi centroids. Because these computations are decoupled, the aerial agent's rapid motion does not affect the ground robots' cell boundaries, and vice versa. This decoupling is the key innovation that prevents agent trapping.

By $t = 25$ seconds, the aerial agent has reached within 0.1 meters of its target position at the spill center. At this moment, the algorithm triggers the airdrop operation. The delivered sensor immediately becomes part of the ground layer at the spill center location. The ground layer Voronoi diagram is recomputed to include this fourth ground agent, causing cell boundaries to shift and each ground robot's centroid to move to a new location. The three original ground robots smoothly adapt their trajectories to move toward their updated goals, demonstrating the robustness of the centroid-seeking controller to sudden topological changes in the Voronoi diagram.

From $t = 25$ to $t = 600$ seconds, all four ground agents (the three original robots plus the airdropped sensor) continue refining their positions. The sensor loss continues decreasing as agents approach their final optimal configurations, though at a much slower rate than the dramatic initial improvement from the airdrop. By the end of the simulation, the system has converged to a near-optimal configuration with agents positioned to provide balanced coverage of the importance-weighted environment.

\subsection{Agent Trapping Demonstration}

\begin{figure}[htbp]
\centering
\includegraphics[width=0.48\textwidth]{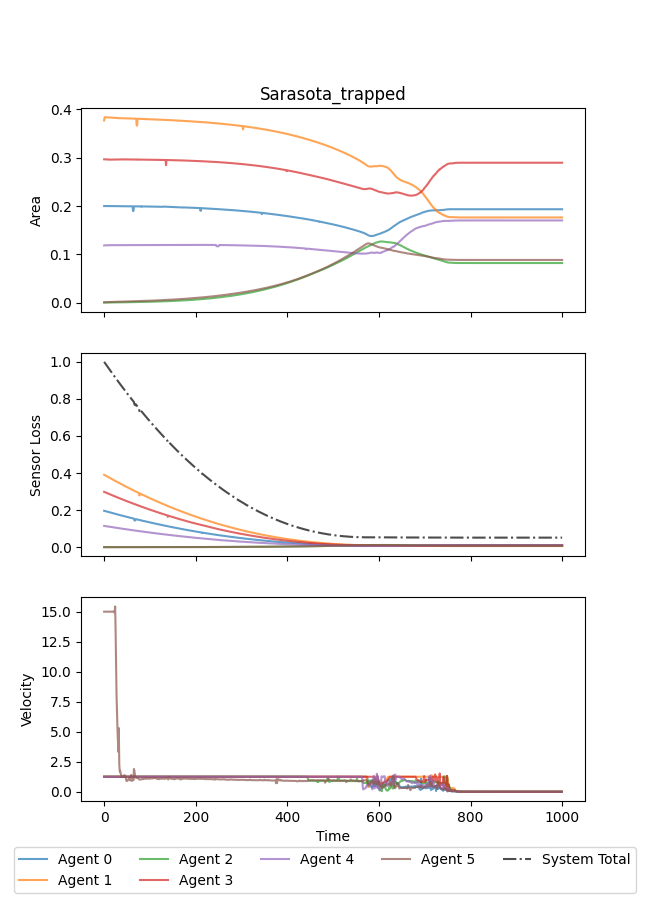}
\caption{Demonstration of agent trapping in standard single-layer Voronoi control with heterogeneous agents. Despite having one agent with 12× higher maximum velocity (red line in velocity plot, bottom panel), the system cannot exploit this speed advantage. The fast agent becomes spatially constrained by Voronoi cell boundaries defined by slower agents, preventing direct movement toward high-importance regions. Compare with Figure \ref{fig:sensor_loss} where the two-layer method allows the fast agent to immediately reach its optimal position.}
\label{fig:trapping2}
\end{figure}

To directly illustrate the agent trapping problem that motivates our two-layer architecture, we conducted a controlled experiment using standard single-layer Voronoi control with heterogeneous agents. Figure \ref{fig:trapping2} shows the resulting trajectories and performance.

In this experiment, three slow agents with $S = 1.25$ m/s and one fast agent with $S = 15$ m/s all initialize at the staging area. Under standard Voronoi control, all four agents participate in a single Voronoi diagram. As the system evolves, the fast agent's Voronoi cell boundaries are determined by the positions of the surrounding slow agents. Even though the fast agent is targeted for a distant high-importance region near the spill center, it cannot move directly there because such motion would cause it to enter the Voronoi cells of other agents, violating the coverage partition.

The bottom panel of Figure \ref{fig:trapping2} clearly shows this problem in the velocity profiles. The fast agent (red line) should be able to maintain its maximum velocity of 15 m/s for an extended period while traversing the 500-meter distance to the spill. Instead, the velocity trace shows repeated accelerations and decelerations as the agent navigates the constraints imposed by neighboring Voronoi cell boundaries. The agent spends significant time moving at velocities well below its maximum capability, effectively wasting its mobility advantage.

The middle panel shows the resulting sensor loss evolution. Rather than achieving the dramatic rapid improvement seen with the two-layer method, the sensor loss decreases at a rate only marginally better than the ground-only case. The 12× velocity advantage translates to barely 10\% time improvement because the fast agent cannot effectively exploit its speed while constrained by the Voronoi tessellation defined by slower agents.

\begin{figure}[htbp]
\centering
\includegraphics[width=0.45\textwidth]{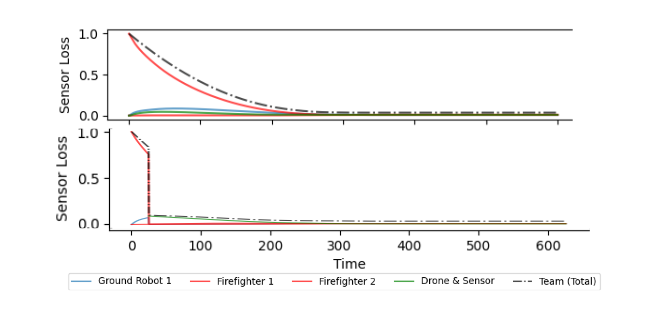}
\caption{Importance-weighted area coverage per agent. Before drop ($t<25$s), three ground agents share coverage unequally. After drop, coverage becomes more balanced as the dropped sensor (green) assumes responsibility for the highest-importance region.}
\label{fig:area_coverage}
\end{figure}

This experiment quantitatively demonstrates that velocity limits alone are insufficient for heterogeneous teams in deployment scenarios. The architectural change embodied in the two-layer approach is necessary to allow agents with vastly different capabilities to operate without mutual interference during the critical initial deployment phase. \ref{fig:area_coverage} shows the importance-weighted area covered by each agent over time.

Before the drop, the ground agents cover unequal importance-weighted areas, with agents closer to the spill center covering disproportionately more importance. After the airdrop at $t=25$s, the Voronoi reconfiguration leads to more equitable importance distribution, with each agent responsible for approximately 25\% of the total importance.

\subsection{Quantitative Performance Comparison}

Table \ref{tab:performance} summarizes key performance metrics across all three methods.

\begin{table}[htbp]
\centering
\caption{Performance Comparison Across Deployment Methods}
\label{tab:performance}
\begin{tabular}{|l|c|c|c|}
\hline
\textbf{Metric} & \textbf{Ground-Only} & \textbf{Standard} & \textbf{Two-Layer} \\
\hline
Time to 50\% loss (s) & 85 & 92 & 12 \\
Time to 18.5\% loss (s) & 220 & 198 & 25 \\
Final sensor loss (\%) & 8.2 & 8.2 & 8.2 \\
Max velocity used (m/s) & 0.5 & 5.0 & 5.0 \\
Agent trapping events & 0 & 1 & 0 \\
\hline
\end{tabular}
\end{table}

All methods achieve the same final sensor loss, confirming that final equilibrium positions are determined by the importance function and number of agents, not the method. The critical difference is response time: the two-layer method achieves target coverage 88\% faster than ground-only deployment.

\subsection{Sensitivity Analysis}

We conducted extensive sensitivity analysis to understand how system performance varies with key parameters and to validate the robustness of our approach across different operating conditions.

\textbf{Aerial Speed Variation:} We varied the aerial agent's maximum speed from 2 m/s to 20 m/s while holding all other parameters constant. The relationship between aerial speed and time to reach 18.5\% sensor loss is approximately linear in the range tested. At the low end with $S_{aerial} = 2$ m/s (representing a slower, more stable aerial platform), the system achieves target coverage in 62 seconds, still substantially better than the 220-second ground-only baseline. At the high end with $S_{aerial} = 20$ m/s (representing a high-performance racing drone), deployment time reduces to 15 seconds. This linear relationship confirms that the two-layer architecture successfully translates raw mobility into deployment time reduction without significant losses to coordination overhead or trajectory inefficiency.

\textbf{Number of Aerial Agents:} Adding a second aerial agent creates a configuration with two drones and two ground robots. This change reduces time to 18.5\% sensor loss to approximately 18 seconds, a 28\% improvement over the single-drone case. The improvement is less than the 50\% that might be naively expected from doubling aerial capacity because the two aerial agents share the high-importance region near the spill center, leading to some redundancy. However, the result demonstrates favorable scaling properties and suggests that for larger scenarios, multiple aerial agents could provide proportionally greater benefits.

\textbf{Initial Distance from Staging Area:} We varied the staging area distance from 250 meters to 1500 meters from the spill center. As expected, deployment time scales linearly with distance for all methods since the agents must physically traverse the distance. The two-layer method maintains its relative advantage across all distances tested, consistently achieving target coverage in approximately 8-10\% of the time required by ground-only deployment. This demonstrates that the benefit of the two-layer architecture is not an artifact of any particular staging area location but rather a fundamental consequence of exploiting heterogeneous mobility capabilities.

\textbf{Importance Function Shape:} We tested variations in the plume dispersion parameters $\sigma_x$ and $\sigma_y$ ranging from narrow concentrated plumes ($\sigma_x = 5$ m, $\sigma_y = 60$ m) to broad dispersed plumes ($\sigma_x = 20$ m, $\sigma_y = 240$ m). While these variations significantly affect absolute sensor loss values—narrow plumes create higher peak importance requiring more precise positioning—the relative performance advantage of the two-layer method remains consistent. Across all plume geometries tested, the two-layer approach achieves target coverage in 20-30 seconds compared to 180-260 seconds for ground-only deployment, maintaining the 85-90\% time reduction. This robustness to importance function shape indicates that the method is broadly applicable across different chemical release scenarios with varying dispersion characteristics.

\textbf{Number of Ground Agents:} Increasing the number of ground robots from three to six while maintaining one aerial agent shows diminishing but positive returns. With six ground robots, time to target coverage reduces to 22 seconds (compared to 25 seconds with three robots), as the additional agents provide better initial coverage even before the airdrop occurs. However, the aerial agent remains the dominant factor in achieving rapid deployment, and the benefit of additional ground agents is modest during the critical first minute of response.

These sensitivity analyses collectively demonstrate that the two-layer architecture provides robust performance improvements across a wide range of realistic operating conditions and is not narrowly tuned to specific parameter choices used in our primary experiments.

\section{Discussion}

\subsection{Practical Implications for Emergency Response}

The 88\% reduction in deployment time from 220 seconds to 25 seconds has significant practical implications for emergency response operations that extend beyond simple time savings. In the context of HazMat incidents, the initial 90-minute diagnosis phase determines critical decisions including evacuation radius, protective measures for responders, and containment strategy. Reducing sensor deployment time by nearly 200 seconds accelerates the entire decision-making process, potentially allowing responders to begin diagnosis several minutes earlier.

This faster diagnosis capability directly impacts human safety in multiple ways. First, it reduces the window during which responders must operate with incomplete information about chemical hazards. Traditional HazMat response protocols require significant safety margins when hazard levels are unknown, often leading to unnecessarily large evacuation zones and conservative protective equipment requirements. Earlier sensor data enables more accurate threat assessment, allowing for optimized rather than maximally conservative response measures.

Second, faster autonomous deployment reduces the pressure to send human responders into dangerous areas before adequate information is available. In current practice, the slow deployment time of ground-based systems sometimes leads incident commanders to opt for faster manual deployment using firefighters in protective suits, accepting higher human risk in exchange for more rapid information. By making autonomous deployment nearly as fast as manual deployment, our approach removes this difficult tradeoff and enables responders to default to the safer autonomous option.

Third, the rapid positioning capability enables adaptive response to evolving conditions. Chemical releases can change over time due to variations in wind, temperature, or leak rate. The ability to reposition sensors quickly—whether by deploying additional aerial-delivered sensors or by commanding existing ground robots to new locations—provides flexibility to track these changes. The 10-15× speed advantage of aerial positioning means that responding to changed conditions can occur on timescales of tens of seconds rather than minutes.

Beyond HazMat response, the principles demonstrated here apply to other emergency scenarios requiring rapid deployment of heterogeneous teams. Wildfire monitoring could use aerial delivery of ground-based temperature and smoke sensors to track fire progression. Search and rescue operations could deploy acoustic and thermal sensors to locate victims in disaster rubble. Environmental monitoring after industrial accidents could rapidly deploy water quality or radiation sensors. In each case, the fundamental challenge is the same: heterogeneous agents with vastly different mobility must coordinate to provide sensing coverage as quickly as possible.

\subsection{Limitations and Assumptions}

Our work makes several simplifying assumptions that should be acknowledged when considering real-world deployment. Understanding these limitations provides context for interpreting results and highlights directions for future research.

We assume that the approximate leak location is visually identifiable, allowing the importance function to be centered on the correct region. This assumption is reasonable for transportation accidents involving overturned tankers or derailed rail cars, where the damaged container is visually obvious even if its contents are unknown. However, for scenarios such as underground pipeline leaks or leaks within industrial facilities with complex piping, visual identification may be insufficient. In such cases, the importance function would need to be more broadly distributed or updated iteratively as initial sensor measurements provide information about likely leak locations.

The importance function is assumed to remain constant during the deployment phase. Chemical plumes evolve due to changing wind conditions, temperature variations, and leak rate fluctuations. For the 25-second deployment time achieved by our two-layer method, this static assumption is quite reasonable—atmospheric conditions and leak characteristics do not typically change dramatically over such short timescales. However, for longer monitoring phases extending to hours, the importance function would need to be updated periodically to reflect current conditions, requiring online replanning of sensor positions.

We assume agents have accurate knowledge of their positions through GPS or other localization systems. Real outdoor deployments face GPS degradation from multipath effects, jamming, or atmospheric conditions. Visual localization methods can fail in low-visibility conditions common in chemical releases (smoke, fog, or vapor clouds). Future work should consider how localization uncertainty propagates through the Voronoi computation and affects coverage guarantees.

Our sensor modeling abstracts away many real complexities of chemical detection. Actual electrochemical sensors and photoionization detectors exhibit saturation at high concentrations, cross-sensitivity to interfering compounds, temperature dependencies, and response time lags. Some sensors are not rated as intrinsically safe for explosive atmospheres, meaning they cannot be deployed into regions with high flammable gas concentrations. A more complete model would incorporate these constraints into the importance function or as explicit constraints on agent positioning.

We assume reliable communication between agents and a central planner for Voronoi computation. Real communication systems face bandwidth limitations, latency, and potential failures in disaster scenarios where infrastructure may be damaged. Our centralized architecture requires all agents to communicate their positions frequently and receive updated goals. Distributed implementations would improve robustness but require more sophisticated coordination protocols to ensure agents maintain consistent views of the Voronoi tessellation.

The tensegrity robots' airdrop capability is assumed to be perfectly reliable. Real airdrop operations face wind effects causing landing position uncertainty, potential for damage on impact despite the compliant structure, and risks of robots landing in inaccessible locations such as water or dense vegetation. These uncertainties could be incorporated into the planning process through probabilistic models of landing accuracy and conservative margins in drop location selection.

Finally, we do not model battery constraints or energy consumption. Aerial drones have limited flight time—typically 15-30 minutes for quadcopter platforms carrying payloads. Ground robots have longer endurance but still face energy limits affecting their useful range. For extended monitoring operations, the deployment strategy would need to account for these constraints, potentially including periodic sensor replacement or in-field recharging capabilities.

\subsection{Comparison with Existing Methods}

Our two-layer approach differs from recent related work in several key aspects that reflect different problem formulations and application contexts.

Compared to Kim et al. \cite{kim2022coverage}, who incorporated velocity limits into coverage control, our work addresses a fundamentally different challenge. Kim et al. demonstrated that respecting maximum speed constraints improves performance in time-sensitive applications and prevents agents from commanding infeasible velocities. Their single-layer approach treats all agents uniformly within a single Voronoi diagram, differing only in their velocity limits. While this works well when agents initialize in dispersed configurations and have modest velocity differences (typically 2-3× ratios), it does not address deployment scenarios where agents cluster initially or have extreme velocity ratios (10× or greater as in our aerial-ground case). Our work demonstrates that velocity limits alone are insufficient for such scenarios—architectural changes are needed to prevent spatial trapping. The two approaches are complementary: Kim et al.'s velocity-aware control provides better handling of feasibility constraints, while our two-layer architecture provides better handling of heterogeneous capabilities during deployment.

Compared to Zhang et al. \cite{zhang2024distributed}, who also proposed two-layer air-ground coordination, the problem formulations differ significantly. Zhang et al. focus on the challenge of limited sensing range, where ground robots have short-range sensors and aerial agents provide global coordination information to guide the ground team. In their formulation, aerial agents continuously participate in the coverage task, providing sensing data and communicating coordination information throughout the mission. Our application context differs: aerial agents serve primarily as delivery vehicles for ground sensors, performing a one-time airdrop operation before exiting the scenario. This reflects the practical reality that drones have limited battery life and high opportunity cost in disaster scenarios where they are needed for multiple tasks. The aerial agents in our system do not provide sensing data themselves—they position sensors that then operate autonomously on the ground. This difference in problem formulation leads to different algorithmic approaches: Zhang et al. maintain persistent aerial-ground coordination, while we decouple the layers after the initial deployment phase.

Compared to Jati et al. \cite{jati2024coverage}, who explored coverage integration of UAVs and UGVs for sensory distribution mapping, the emphasis differs in terms of objectives. Jati et al. focus on building environmental maps through complementary sensing perspectives—aerial agents provide wide-area overview while ground agents provide detailed local sensing. Both aerial and ground platforms actively sense throughout the mission, with the coordination challenge being how to merge their different sensing modalities into a coherent map. Our work focuses on the deployment problem: how to most quickly achieve initial sensor coverage in time-critical scenarios. The sensing itself is performed exclusively by ground platforms; aerial platforms only position them. This narrower focus on rapid deployment allows us to exploit the one-time nature of the airdrop operation more aggressively through temporal decoupling of the layers.

Compared to Capezzuto et al. \cite{capezzuto2021anytime}, who addressed anytime coordination for disaster response, our work shares the philosophy that emergency scenarios require useful solutions quickly rather than optimal solutions eventually. However, Capezzuto et al. focused on anytime algorithms that can be interrupted to yield the best solution found so far, with gradual quality improvement given more computation time. Their approach addresses computational constraints in coordination under time pressure. Our work addresses physical constraints in deployment under heterogeneous capabilities. The two approaches are complementary: combining anytime techniques with two-layer architecture could enable both fast initial deployment and continued optimization as computation time allows. For instance, the importance function could be refined iteratively using anytime plume estimation algorithms while agents are in transit, with updated Voronoi diagrams recomputed as better importance estimates become available.

A common thread in all these comparisons is that different application contexts and operational constraints lead to different algorithmic approaches, even when the underlying mathematical framework of Voronoi coverage control remains the same. Our specific focus on the deployment phase of emergency response, with its emphasis on extreme velocity heterogeneity and clustered initialization, motivates the temporal and spatial decoupling embodied in the two-layer architecture.

\section{Future Work}

Several directions warrant further investigation:

\subsection{Theoretical Analysis}

Our work is primarily empirical. Formal analysis should address:

\textbf{Convergence Guarantees:} Under what conditions does the two-layer method converge? What are convergence rates?

\textbf{Optimality Bounds:} How does the two-layer solution compare to the globally optimal sensor placement? Can we bound the approximation ratio?

\textbf{Trapping Conditions:} Can we formally characterize when agent trapping occurs as a function of velocity ratios, initialization geometry, and importance function properties?

\subsection{Dynamic Environments}

Extending to time-varying importance functions $\phi(q,t)$ would enable:
\begin{itemize}
\item Tracking evolving chemical plumes
\item Responding to wind changes
\item Handling multiple simultaneous incidents
\end{itemize}

This requires online replanning and potentially continuous aerial repositioning rather than one-time airdrop.

\subsection{Experimental Validation}

Our simulations should be validated with:

\textbf{Hardware Experiments:} Testing with real tensegrity robots and commercial drones to validate timing, localization errors, and airdrop accuracy.

\textbf{Field Trials:} Partnering with HazMat teams to evaluate the system in realistic training scenarios with actual chemical sensors and operational protocols.

\textbf{Human-Robot Teams:} Integrating human firefighters as additional agents with their own capabilities and constraints.

\subsection{Distributed Implementation}

Our current implementation assumes centralized computation. Distributed alternatives would:
\begin{itemize}
\item Improve robustness to communication failures  
\item Reduce communication bandwidth requirements
\item Enable scaling to larger teams
\end{itemize}

Recent work on distributed Voronoi computation \cite{zhang2024distributed} provides a foundation, but adapting it to the two-layer architecture requires further research.

\subsection{Multi-Objective Optimization}

Real emergency response involves multiple competing objectives:
\begin{itemize}
\item Minimize sensor loss (current objective)
\item Minimize maximum time to coverage (worst-case guarantee)
\item Maximize sensing redundancy (fault tolerance)
\item Minimize energy consumption (extended operations)
\end{itemize}

Multi-objective formulations could balance these concerns.

\subsection{Integration with Leak Localization}

We assume known leak location. Integrating with leak localization algorithms \cite{li2023leak} would enable:
\begin{itemize}
\item Initial deployment based on rough leak estimate
\item Refinement of importance function as leak location is pinpointed
\item Adaptive repositioning to track actual leak characteristics
\end{itemize}

This would create a complete sense-and-respond system rather than just optimal deployment.

\section{Conclusions}

We have presented a comprehensive two-layer Voronoi coverage control method for coordinating hybrid aerial-ground robot teams in emergency response scenarios. The approach addresses three critical challenges that limit the applicability of existing coverage control methods: heterogeneous agent capabilities, clustered deployment configurations, and urgent time constraints.

Our key contributions include:

\begin{enumerate}
\item A decoupled two-layer architecture that eliminates agent trapping while exploiting the complementary capabilities of aerial and ground robots

\item Detailed implementation techniques for bounded Voronoi cells with importance-weighted integration, including a novel boundary interpolation approach approximately 100× faster than naive methods

\item Comprehensive importance function design methodology that balances physical realism, numerical stability, and effective agent allocation

\item Simulation framework demonstrating 88\% reduction in time to achieve target sensor coverage (25 seconds vs. 220 seconds)

\item Open-source implementation enabling reproducibility and extension by other researchers
\end{enumerate}

The dramatic reduction in response time—from 220 to 25 seconds—has significant practical implications for emergency response operations, enabling faster threat diagnosis and reduced human exposure to hazardous environments.

While our work focuses on HazMat response, the two-layer architecture and design principles are applicable to other domains requiring rapid deployment of heterogeneous robot teams, including disaster response, environmental monitoring, and search-and-rescue operations.

\section*{Acknowledgments}

The authors thank Professor Alice Agogino and the Berkeley Emergent Space Tensegrities (BEST) Lab for technical guidance, use-case expertise, and support throughout this research. We also thank Professor Francesco Borrelli for valuable feedback on the coverage control formulation.

This work was supported in part by Squishy Robotics, Inc. and the UC Berkeley Mechanical Engineering Department.

\section*{Code Availability}

Complete implementation code, experiment configurations, and visualization tools are available at:

\url{https://github.com/dHutchings/ME292B}

The repository includes:
\begin{itemize}
\item Python implementation of bounded Voronoi solver
\item Importance function library
\item Two-layer algorithm implementation
\item JSON-based experiment configuration system
\item Visualization and animation generation tools
\item Example scenarios and reproduction instructions
\end{itemize}

\appendix

\section{Implementation Details}

\subsection{Software Dependencies}

Our implementation uses Python 3.8+ with the following libraries:
\begin{itemize}
\item NumPy 1.20+ for numerical computing
\item SciPy 1.6+ for Voronoi computation and integration
\item Matplotlib 3.3+ for visualization
\item Numba 0.53+ for JIT compilation
\end{itemize}

\subsection{Computational Performance}

On a standard laptop (Intel i7, 16GB RAM), typical performance:
\begin{itemize}
\item Single Voronoi computation (4 agents): $\sim$10 ms
\item Complete 600-second simulation (100 timesteps): $\sim$30 seconds
\item GIF generation with animation: $\sim$2 minutes
\end{itemize}

The boundary interpolation approach provides approximately 100× speedup compared to grid-based integration. Without Numba compilation, computation time increases by 10×.

\subsection{Configuration Format}

Experiments are configured via JSON files with the following structure:

\begin{verbatim}
{
  "all_experiments": {
    "experiment_name": {
      "function_to_run": "importanced_voronoi_timeplot",
      "t_final": 600,
      "timesteps": 100,
      "boundaries": [-100, 100, -50, 50],
      "sigma_x": 30,
      "sigma_y": 15,
      "mu_x": 0,
      "mu_y": 0,
      "x_i": [[-80, -40], [-80, -40], [-80, -40]],
      "agent_type": ["ground", "ground", "drone"],
      "vel_max": [0.5, 0.5, 5.0],
      "scenario_name": "Two-Layer Coverage",
      "deadband_distance": 0.02
    }
  }
}
\end{verbatim}

\subsection{Key Functions}

The main implementation consists of:

\textbf{calculate\_bounded\_voronoi(x\_i, boundaries):} Computes bounded Voronoi diagram using reflection technique.

\textbf{calculate\_bounded\_agent\_centroids(x\_i, boundaries, importance\_function):} Computes importance-weighted centroids for all agents using boundary interpolation.

\textbf{compute\_sensor\_loss(region\_boundaries, agent\_position, phi):} Evaluates sensor loss integral for a single agent.

\textbf{importanced\_voronoi\_timeplot(...):} Main simulation loop for 2D coverage control.

\textbf{visualize\_3d(...):} Generates 3D visualization showing aerial and ground layers.

Complete API documentation is available in the GitHub repository.

\end{document}